\documentclass[runningheads]{llncs}
\usepackage{graphicx}
\usepackage{multirow}

\begin{document}
\title{PC-U Net: Learning  to Jointly Reconstruct and Segment the Cardiac Walls in 3D from CT Data}

\titlerunning{PC-U Net}

\author{Meng Ye\inst{1} \and Qiaoying Huang\inst{1} \and
Dong Yang\inst{2} \and Pengxiang Wu\inst{1} \and Jingru Yi\inst{1} \and Leon Axel\inst{3} \and Dimitris Metaxas\inst{1}}

\authorrunning{Ye et al.}

\institute{Department of Computer Science, Rutgers University, Piscataway, NJ 08854, USA 
\email{my389@cs.rutgers.edu} \and Nvidia Corporation, Bethesda, MD 20814, USA \and Department of Radiology, New York University, New York, NY 10016, USA}

\maketitle              
\begin{abstract}
The 3D volumetric shape of the heart's left ventricle (LV) myocardium (MYO) wall provides important information for diagnosis of cardiac disease and invasive procedure navigation. 
Many cardiac image segmentation methods have relied on detection of region-of-interest as a pre-requisite for shape segmentation and modeling. 
With segmentation results, a 3D surface mesh and a corresponding point cloud of the segmented cardiac volume can be reconstructed for further analyses. 
Although state-of-the-art methods (e.g., U-Net) have achieved decent performance on  cardiac image segmentation in terms of accuracy, these segmentation results can still suffer from imaging artifacts and noise, which will lead to inaccurate shape modeling results. 
In this paper, we propose a PC-U net that jointly reconstructs the point cloud of the LV MYO wall directly from volumes of 2D CT slices and generates its segmentation masks from the predicted 3D point cloud. Extensive experimental results show that by incorporating a shape prior from the point cloud, the segmentation masks are more accurate than the state-of-the-art U-Net results in terms of Dice's coefficient and Hausdorff distance. The proposed joint learning framework of our PC-U net is beneficial for automatic cardiac image analysis tasks because it can obtain simultaneously the 3D shape and segmentation of the LV MYO walls.

\keywords{Point cloud  \and Segmentation \and LV shape modeling.}
\end{abstract}
\section{Introduction}
\begin{figure}[!t]
\centering
\includegraphics[width=0.7\textwidth]{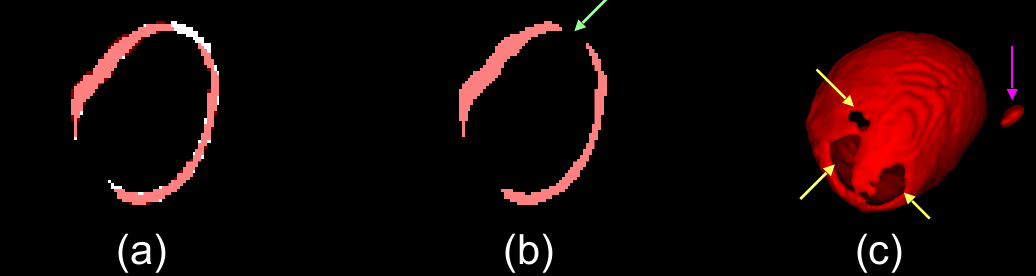}
\caption{The shape generated from U-Net based segmentation results can be erroneous (holes indicated by yellow arrows) due to the thin wall structure of the myocardium apex (green arrow). (a) is the ground truth (in white color) of segmentation overlaid by the U-Net segmentation result shown in (b). (c) is the LV MYO wall shape reconstructed from (b).
}
\label{fig1}
\end{figure}

Cardiovascular disease is one of the major causes of human death worldwide. The shape of the left ventricle (LV) myocardium (MYO) wall plays an important role in diagnosis of cardiac disease, and in surgical and other invasive cardiac procedures~\cite{medrano2015statistical}.
It allows the subsequent cardiac LV motion function analysis to determine the presence and location of possible cardiac disease~\cite{juergens2004multi}. 
In order to automatically evaluate cardiac function measures such as the ejection fraction (EF), most previous methods first segment the LV blood pool or LV MYO from images~\cite{lopez2015three} then generate a shape model from the segmentation results.
However, this sequential pipeline is not ideal since the shape reconstruction strongly depends on the precision of the segmentation algorithms. 
As the example shown in Fig.~\ref{fig1}, due to the imperfect segmentation result, the generated 3D shape is erroneous and not suitable in a practical clinical scenario. 
It is therefore imperative to develop a more robust approach for LV MYO segmentation jointly with accurate and clinically useful myocardium 3D surface reconstruction.

 The 3D shape of an object can be described as a 3D mesh or point cloud (PC). A 3D mesh contains a set of vertices, edges and faces, which can be represented as a graph $M=(V,E)$, where $M$ is the mesh, $V=\left \{ v_{i} \right \}_{i=1}^{N}$ is the set of $N$ vertices and $E=\left \{ e_{j} \right \}_{j=1}^{K}$ is a set of $K$ edges with each connecting two vertices. Each vertex $v$ has a 3D coordinate $(x, y, z)$. The set of the vertices $V$ forms a point cloud. Point cloud-based shape representations have received increasing attention in medical image analysis. In~\cite{balsiger2019learning}, a convolutional neural network (CNN)-based classifier was first trained to estimate a point cloud from 3D volumetric images. In particular, a point cloud network was trained to classify each point into foreground or background in order to refine the segmentation mask of the peripheral nerve. In~\cite{cai2019end}, an organ point-network was introduced to learn the shape of abdomen organs which was described as points located on the organ surface. This point-network was trained together with a CNN-based segmentation model to help improving organ segmentation results in a multi-task learning manner. Recently, another method, named PointOutNet was proposed to predict a 3D point cloud from a single RGB image~\cite{fan2017point}. In~\cite{zhou2019one}, the same model was applied to conduct one-stage shape instantiation for the right ventricle (RV) from a single 2D long-axis cardiac MR image. However, their method required an optimum imaging plane, which was decided by a predefined 3D imaging process to acquire a synchronized 2D image as the input to the PointOutNet.

In this paper, we propose a PC-U net to reconstruct a 3D point cloud of the LV MYO shape from a volume of CT images. In order to segment the myocardium wall for other applications such as global heart function evaluation, we design this PC-U net to reconstruct the slice segmentation masks from the point cloud directly. 
The benefits of our PC-U net are twofold. First, the reconstructed  LV MYO wall 3D shape is no longer dependent  on the precision of the segmentation results. Second, the LV MYO wall segmentation results are reconstructed from the point cloud, therefore it combines the reconstructed shape prior into the segmentation branch of the network, which, in turn, makes the segmentation results more accurate~\cite{painchaud2019cardiac,yue2019cardiac}. To the best of our knowledge, this is the first work that explicitly utilizes features from an estimated shape prior of an organ in the form of a point cloud  to reconstruct segmentation masks.  

\section{Proposed Method}
\subsection{PC-U Net}
\begin{figure}[!t]
\includegraphics[width=\textwidth]{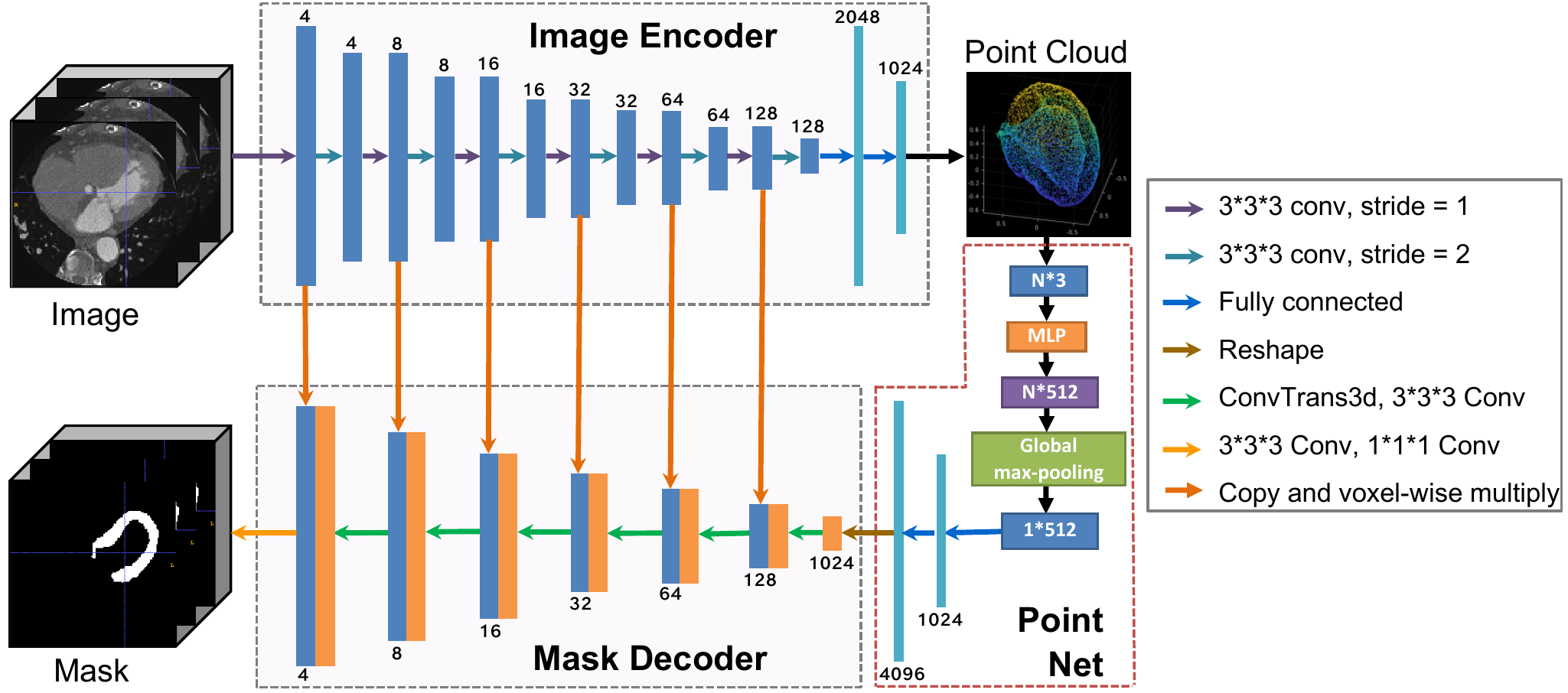}
\caption{An overview of the proposed PC-U net for 3D point cloud reconstruction and segmentation of the LV MYO wall simultaneously (best view in color). 
The numbers on the top or bottom of a bar depict the number of output channels for convolutional layers or fully connected layers. $N$ is the point number of a point cloud.} \label{fig2}
\end{figure}

Fig.~\ref{fig2} shows the architecture of our proposed PC-U net for 3D point cloud reconstruction and segmentation of the LV MYO wall. The PC-U net consists of three parts: image encoder, point net, and mask decoder.

\subsubsection{Image Encoder} As shown in Fig.~\ref{fig2}, the image encoder takes 3D volumes of CT images as input and extracts image features $IF(x',y',z')$ at each voxel $(x',y',z')$. This branch is used for 3D point cloud reconstruction, which is similar to the PointOutNet~\cite{zhou2019one}, except that the inputs are 3D volumetric images and convolution kernels are 3D.

\subsubsection{Point Net} The point net consists of a multi-layer perceptron (MLP), a global max-pooling layer and two fully connected layers. The MLP is consisted of three convolutional layers (with $32$, $128$, $512$ channels, respectively), followed by ReLU activation. This branch is used to extract point features $PF(x, y, z)$ at each 3D point $(x, y, z)$ of the point cloud~\cite{qi2017pointnet}. These point features will be used to reconstruct segmentation masks. In this way, the point cloud can provide a shape prior for the segmentation task explicitly.

\subsubsection{Mask Decoder} The lower branch is the mask decoder, which is used for LV MYO wall segmentation. We refer the process of reconstructing dense segmentation masks from a sparse point cloud as segmentation.
It consists of up-sampling layers and convolutional layers.
In order to take advantage of the more detailed contextual image features in the shallow layers, we use skip connections between the upper branch and the lower branch. 
This fashion makes the upper and lower branches similar to the encoder and decoder of the U-Net~\cite{ronneberger2015u}. 
In particular, the features from the encoder are multiplied by the features in the decoder voxel-wisely. 
We empirically found that voxel-wise multiplication operation could make the output masks have less outliers.
The significant difference between the proposed PC-U net and a U-Net is that PC-U net allows the intermediate output of the image encoder be the point cloud, which serves as an effective regularizer and provides a shape prior for segmentation. 
Given the point features $PF(x, y, z)$ and image features $IF(x',y',z')$, the mask decoder branch will learn an implicit function $f$ represented by the up-sampling layers and convolutional layers:
\begin{equation}
f(PF(x,y,z), IF(x',y',z')) = m:m\in \left \{ 0, 1 \right \},
\end{equation}
where $m$ is the binary segmentation mask. 
\subsection{Loss Function}
We use the Chamfer loss \cite{yang2018foldingnet} to measure the distance between the predicted point cloud and the ground-truth point cloud. It can be formulated as:
\begin{equation}
l_{p}=\frac{1}{\left | Y_{P} \right |}\sum_{y_{p}\in Y_{P}}{min_{y_{g}\in Y_{G}}}\left \| y_{p}-y_{g} \right \|_{2}+\frac{1}{\left | Y_{G} \right |}\sum_{y_{g}\in Y_{G}}{min_{y_{p}\in Y_{P}}}\left \| y_{g}-y_{p} \right \|_{2},
\end{equation}
where the first term ${min_{y_{g}\in Y_{G}}}\left \| y_{p}-y_{g} \right \|_{2}$ enforces that any 3D point $y_{p}$ in the predicted point cloud $Y_{P}$ has a matching 3D point $y_{g}$ in the ground-truth point cloud $Y_{G}$, and vice versa for the second term ${min_{y_{p}\in Y_{P}}}\left \| y_{g}-y_{p} \right \|_{2}$.

For segmentation, we adopt the soft Dice loss~\cite{milletari2016v} to measure the error between the predicted mask and the ground truth:
\begin{equation}
l_s=1-\frac{2\sum_{i}^{N}{p_{i}g_{i}}}{\sum_{i}^{N}(p_{i}^{2}+g_{i}^{2})},
\end{equation}
where $N$ is the total number of voxels, $p_{i}$ is the $i^{th}$ voxel of the predicted mask and $g_{i}$ is the $i^{th}$ voxel of the ground-truth mask.

The PC-U net is trained end-to-end and optimized by the following weighted loss of $l_p$ and $l_s$. 
\begin{equation}
Loss=l_p + \lambda \cdot l_s,
\end{equation}
where $\lambda$ is a balancing weight.
We empirically set $\lambda=0.001$ via the grid search.

\section{Experiments}
\begin{figure}[!t]
\centering
\includegraphics[width=0.8\textwidth]{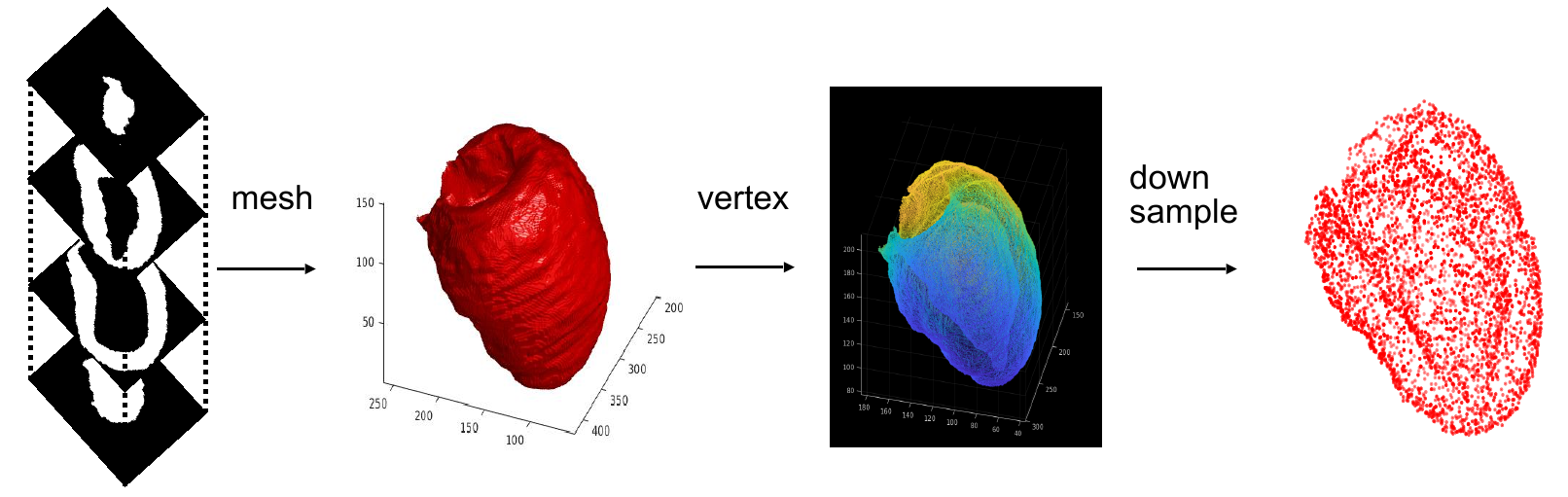}
\caption{The process of how to get the ground-truth point cloud: (1) A mesh is first built from the 3D volumetric masks; (2) The vertices of the mesh form a dense point cloud; (3) We down-sample a sparse point cloud from the dense point cloud.} \label{fig3}
\end{figure}

We validated the proposed PC-U net on a publicly available 3D cardiac CT dataset \cite{zhuang2016multi}, which contains 20 subjects. 
In order to deal with the rather small number of available training samples, we used elastic deformation to augment the training set by 200 times. 
After augmentation, there are 4,000 samples in total.
We first cropped out the region-of-interest of LV and sampled all volumes and their labels to 1.0 $mm$ isotropically. Then the image volume was resized to a fixed size as $(128, 128, 64)$. 
The intensity values of the CT volumes were first divided by 2048 and then clamped between $[-1, 1]$. 
The ground-truth point cloud generation process is shown in Fig.~\ref{fig3}.
All point clouds were centered at $(0, 0, 0)$ by subtracting their own center coordinates. 
In our experiments, the total number of points was set to be $N = 4096$.
To show the robustness and effectiveness of our PC-U net, we performed 4-fold cross validation experiments. 

Since there is no other work that jointly reconstructs a point cloud and segments the LV MYO wall, we will compare the two tasks separately.
For point cloud reconstruction, PointOutNet~\cite{fan2017point} is the baseline model, which takes as input  a single image slice~\cite{zhou2019one} and outputs a 3D point cloud, as shown in the upper encoder branch of Fig.~\ref{fig2}.
Depending on the input data (a single slice of long axis image or the whole 3D volumetric images) and the convolutional kernel (2D or 3D), we create three variants: \textit{PointOutNet-single-slice}, \textit{PointOutNet-volume-2DConv} and \textit{PointOutNet-volume-3DConv}.
For segmentation, the state-of-the-art model U-Net~\cite{payer2017multi} is compared to our method.
We  consider 3D volumetric images as input, giving two models: \textit{UNet-volume-2DConv} and \textit{UNet-volume-3DConv}.
Our \textit{PC-Unet-2DConv} and \textit{PC-Unet-3DConv} also take as input  3D volumetric images, and output the 3D point clouds and segmentation masks simultaneously.
The goal of this experiment is to validate the significance of using 3D input and solving two tasks together as well as show how they benefit each other. 

All models were trained on a single NVIDIA Tesla K40 GPU with Adam optimizer and the learning rate was set at $1e^{-4}$. The Chamfer distance (CD) error between prediction and ground truth was used to measure the point cloud reconstruction performance. 
Dice's coefficient and Hausdorff distance (HD) error were used to evaluate the segmentation performance.

\subsubsection{Experimental Results.} Table 1 shows the average performance on LV MYO wall point cloud reconstruction and segmentation of the 4-fold cross validation. As we can see from the table, with more information contained in a volume input (PointOutNet-single-slice $v.s.$ PointOutNet-volume-2DConv), the reconstructed point cloud gets better in terms of CD error. In addition, 3D convolution outperforms 2D convolution (PointOutNet-volume-2DConv $v.s.$ PointOutNet-volume-3DConv) on the point cloud reconstruction task. PC-Unet-3DConv achieves the best point cloud reconstruction result among all the models, which could be attributed to the joint learning of point cloud reconstruction and segmentation, apart from using a volume input and 3D convolution.

\begin{table*}[!t]
	\centering
    \caption{Average performance of 4-fold cross validation on the point cloud reconstruction and segmentation of left ventricle myocardium (MYO) wall. The mean and standard deviation values over all the folds are reported.}
	\label{tbl:exp2}
	\begin{tabular}{|c|c|cc|}
	\hline
    \multirow{2}{*}{Method} & PC Reconstruction & \multicolumn{2}{c|}{Segmentation} \\
    \cline{2-4} 
     & CD~($mm$) & Dice & HD~($mm$) \\
		\hline
		PointOutNet-single-slice & $1.489\pm0.547$ &- & - \\
        PointOutNet-volume-2DConv & $1.454\pm 0.422$ & - & - \\ 
        PointOutNet-volume-3DConv  & $1.330\pm0.330$ & - & - \\
		UNet-volume-2DConv    & - & $0.843\pm0.024$ & $16.477\pm8.311$ \\
		UNet-volume-3DConv    & - & $0.877\pm0.012$ & $10.446\pm4.489$ \\
		PC-Unet-2DConv & $1.278\pm0.249$ & $0.838\pm0.026$ & $10.894\pm2.176$ \\
		PC-Unet-3DConv (Ours) & $\mathbf{1.276\pm0.168}$ & $\mathbf{0.885\pm0.011}$ & $\mathbf{7.050\pm1.103}$ \\
		\hline
	\end{tabular}
\end{table*}

\begin{figure}[!t]
\centering
\includegraphics[width=0.7\textwidth]{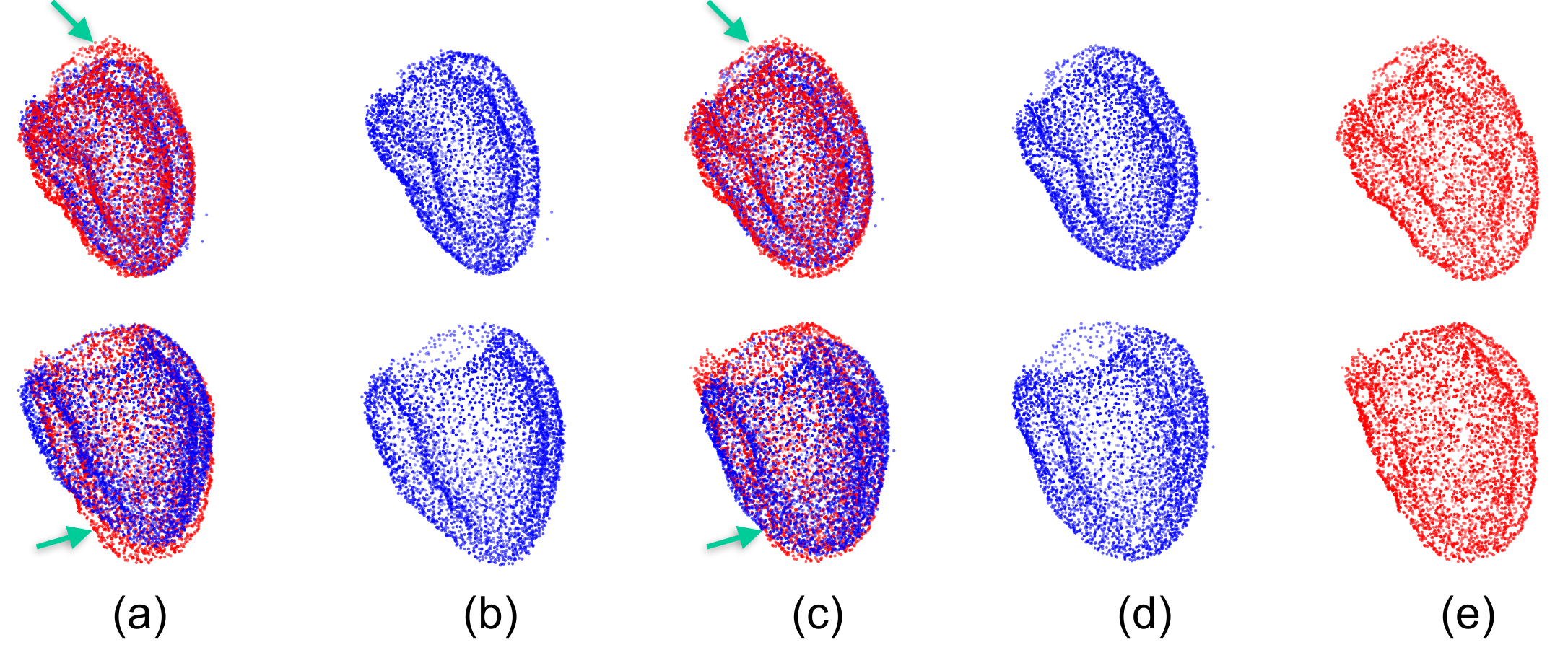}
\caption{LV MYO wall point cloud reconstruction results. (a) PointOutNet (blue) overlaid with ground-truth point cloud (red). (b) PointOutNet. (c) PC-U net (blue) overlaid with ground-truth point cloud (red). (d) PC-U net (Ours). (e) ground-truth point cloud.} \label{fig4}
\end{figure}

Fig.~\ref{fig4} shows two cases (the first and the second rows) of visual comparisons between the baseline (PointOutNet-single-slice) and the proposed model (PC-Unet-3DConv). It is clear that baseline model fails to reconstruct the points on the thin areas of myocardium including the atrioventricular ring and the apex (indicated by arrows). This limitation has been pointed out by the authors in \cite{zhou2019one}. When combining with more 3D contextual information in contrast to a single input 2D image slice, the proposed PC-U net successfully reconstructs the 3D points on these thin regions of myocardium.

\begin{figure}[!t]
\centering
\includegraphics[width=0.9\textwidth]{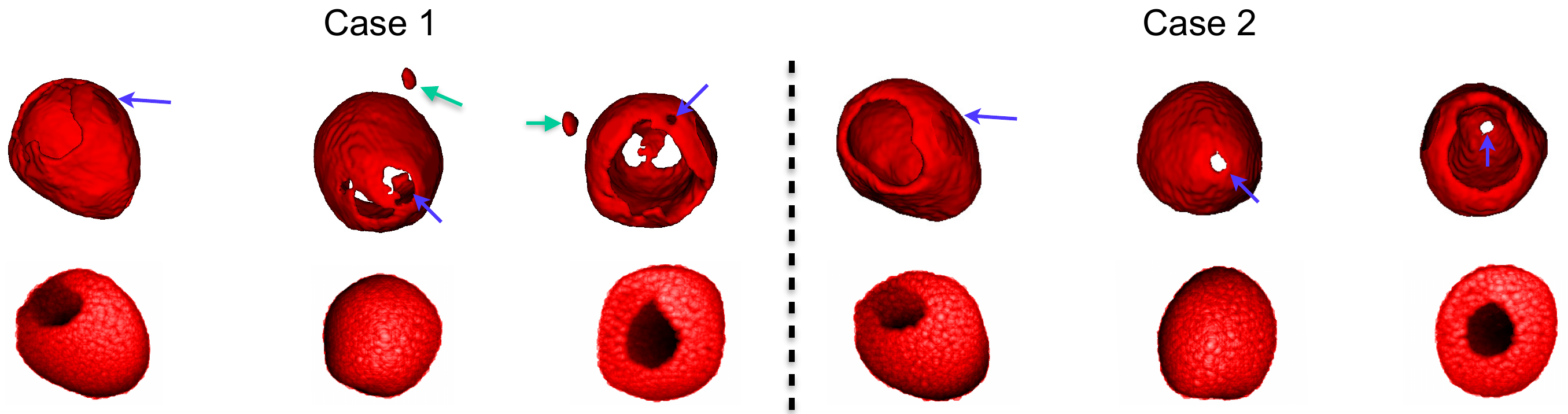}
\caption{LV MYO wall shape modeling from U-Net segmentation (first row) and point cloud predicted by the proposed PC-U net (Ours, second row). Case one: first to third column. Case two: fourth to sixth column. Each column for each case represents a specific view.} \label{fig5}
\end{figure}
For segmentation of the LV MYO, PC-Unet-3DConv achieves a slightly better result in terms of Dice's coefficient compared to UNet-volume-3DConv.
Note that it achieves a significant smaller Hausdorff distance error. 
Again, 3D convolution outperforms 2D convolution (UNet-volume-2DConv $v.s.$ UNet-volume-3DConv,  PC-Unet-2DConv $v.s.$ PC-Unet-3DConv) on the segmentation task. 
Although PC-Unet-2DConv performs slightly worse than UNet-volume-2DConv in terms of Dice's coefficient, its Hausdorff distance error is much smaller than UNet-volume-2DConv, suggesting the significant benefit of using the shape prior provided by the point cloud on the segmentation task.
We qualitatively show two cases of the shape reconstructed from the segmentation results of UNet-volume-3DConv and the predicted point cloud of the proposed PC-Unet-3DConv in Fig.~\ref{fig5}, respectively. 
It is clear that due to error segmentation, there can be holes on the surface of the LV MYO wall shape (indicated by blue arrows). 
When our PC-U net takes as input the 3D volumetric images it successfully predicts a complete shape of the LV MYO wall without any holes.
The large Hausdorff distance error of UNet-volume-3DConv indicates that there are some isolated outliers in the segmentation results, shown as green arrows in Fig.~\ref{fig5}.

\begin{table*}[!t]
	\centering
    \caption{Average performance of the ablation study.}
	\label{tbl:exp2}
	\begin{tabular}{|c|c|cc|}
	\hline
    \multirow{2}{*}{Method} & PC Reconstruction & \multicolumn{2}{c|}{Segmentation} \\
    \cline{2-4} 
     & CD~($mm$) & Dice & HD~($mm$) \\
		\hline
		PC-Mask-Decoder    & - & $0.725$ & $8.682$ \\
		PC-Unet-no-skip & $1.182$ & $0.699$ & $7.896$ \\
		PC-Unet (Ours) & $\mathbf{1.149}$ & $\mathbf{0.884}$ & $\mathbf{5.862}$ \\
		\hline
	\end{tabular}
\end{table*}

\begin{figure}[!t]
\centering
\includegraphics[width=0.8\textwidth]{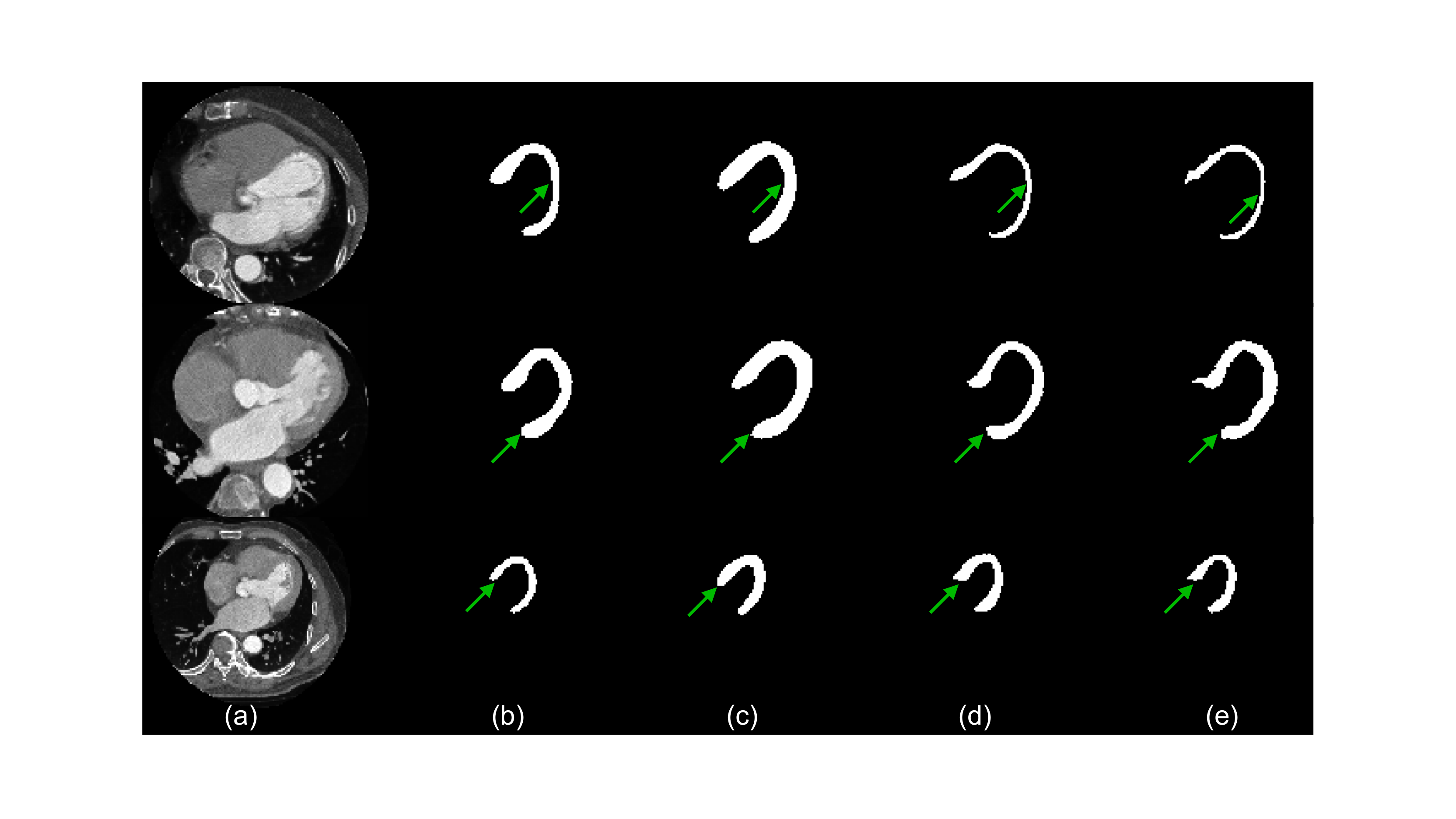}
\caption{LV MYO segmentation results. (a) the CT image slice. (b) PC-Mask-Decoder. (c) PC-Unet-no-skip. (d) PC-U net (Ours). (e) ground-truth mask.} \label{fig6}
\end{figure}

\subsubsection{Ablation Study} 
In this experiment, we demonstrate how the PC-U net can recover segmentation details using skip connections between encoder and decoder. 
We test 3 models here with a 1-fold experiment. 
The first model is composed of the Point Net and Mask Decoder with the ground-truth point cloud as input (PC-Mask-Decoder). 
With this model, we plan to demonstrate that CNNs can reconstruct dense masks from a sparse point cloud. 
The second model is the PC-U net without skip connections (PC-Unet-no-skip). 
The third model is the proposed PC-U net model. 
All the models use 3D convolution. 
The point cloud reconstruction and segmentation results are shown in Table 2.

From Table 2, although the point cloud is a sparse representation of the 3D shape of LV MYO wall, we can use the Point Net to extract point features efficiently with which the Mask Decoder can reconstruct a dense segmentation mask. 
To our knowledge, this is the first work that uses CNNs to reversely reconstruct a dense mask from a sparse point cloud.
However, as shown in Fig.~\ref{fig6}, the PC-Mask-Decoder alone can only reconstruct a coarse mask. 
As indicated by the arrows, some details of the mask cannot be recovered very well. 
The PC-Unet-no-skip model takes input as the volumetric images but it does not consider the detailed contextual image features extracted by the Image Encoder, again resulting in coarse segmentation. 
Another reason for its lower Dice's coefficient compared to PC-Mask-Decoder is that the point cloud reconstructed by the PC-Unet-no-skip model still has some errors, which will introduce bias in point features. 
The proposed PC-U net achieves the best point cloud reconstruction and recovers most details of the segmentation by using skip connections and joint learning.

\section{Conclusion}
In this work, we proposed a PC-U net to reconstruct an LV MYO wall point cloud from 3D volumetric CT images. 
By combining with the 3D context information in the 3D volumetric images, the PC-U net could successfully reconstruct the 3D point cloud of the LV MYO wall. 
Our PC-U net models the shape of the LV MYO wall directly from the 3D volumetric images, which is accurate and avoids errors induced by potentially inaccurate segmentation results. We demonstrated that the PC-U net could not only find the shape of LV MYO wall, but also obtain its segmentation results simultaneously. With a shape prior provided by the point cloud, the segmentation results are more accurate than state-of-the-art U-Net results in terms of Dice's coefficient and Hausdorff distance.

%
%
%
\bibliographystyle{splncs04}
\bibliography{ref}
%




\end{document}